\documentclass[nouppercase]{kaia}
\usepackage{amsmath}
\usepackage{amssymb}
\usepackage{mathtools}
\usepackage{amsthm}
\usepackage{enumitem, setspace, lipsum}
\usepackage[textsize=tiny]{todonotes}
\usepackage{enumitem}
\usepackage{tikz}
\newcommand*\circled[1]{\tikz[baseline=(char.base)]{
            \node[shape=circle,draw,inner sep=2pt] (char) {#1};}}

\title{Explanation on Pretraining Bias of Finetuned Vision Transformer}

\author{Bumjin Park}{a}
\author{Jaesik Choi}{b}
\affiliation{Graduate School of AI, KAIST, Daejeon, Korea, bumjin@kaist.ac.kr}{a}
\affiliation{Graduate School of AI, KAIST, Daejeon, Korea, jaesik.choi@kaist.ac.kr, Corresponding author}{b}

\begin{document}

\maketitle
\begin{abstract}
	As the number of fine tuning of pretrained models increased, understanding the bias of pretrained model is essential. However, there is little tool to analyse transformer architecture and the interpretation of the attention maps is still challenging. To tackle the interpretability, we propose Input-Attribution and Attention Score Vector (IAV) which measures the similarity between attention map and input-attribution and shows the general trend of interpretable attention patterns. We empirically explain the pretraining bias of supervised and unsupervised pretrained ViT models, and show that each head in ViT has a specific range of agreement on the decision of the classification.  We  show that generalization, robustness and entropy of attention maps are not property of pretraining types. On the other hand, IAV trend can separate the pretraining types. 

\end{abstract}

\begin{keywords}
	 Explainable AI, Computer Vision, Deep Learning
\end{keywords}

\section{Introduction}

% ViT
Recently, the applications of pretrained foundation vision models have been increased in several domains. One of the important foundation models is Vision Transformer (ViT)\cite{viT} which opened another paradigm in computer vision by pretraining large amount of images. Unlike Convolution Neural Network (CNN)\cite{krizhevsky2017imagenet_cnn} which is known to have local receptive field with convolutions, ViT has global receptive field by the spatial Multi-head Self-Attention (MSA) mechanism \cite{vaswani2017attention} for image patches. To learn the global receptive field, pretraining of the ViT is essential and without it the model does not generalize well \cite{chen2021empirical}.

\begin{figure}[ht]
\centering
\includegraphics[width=8cm]{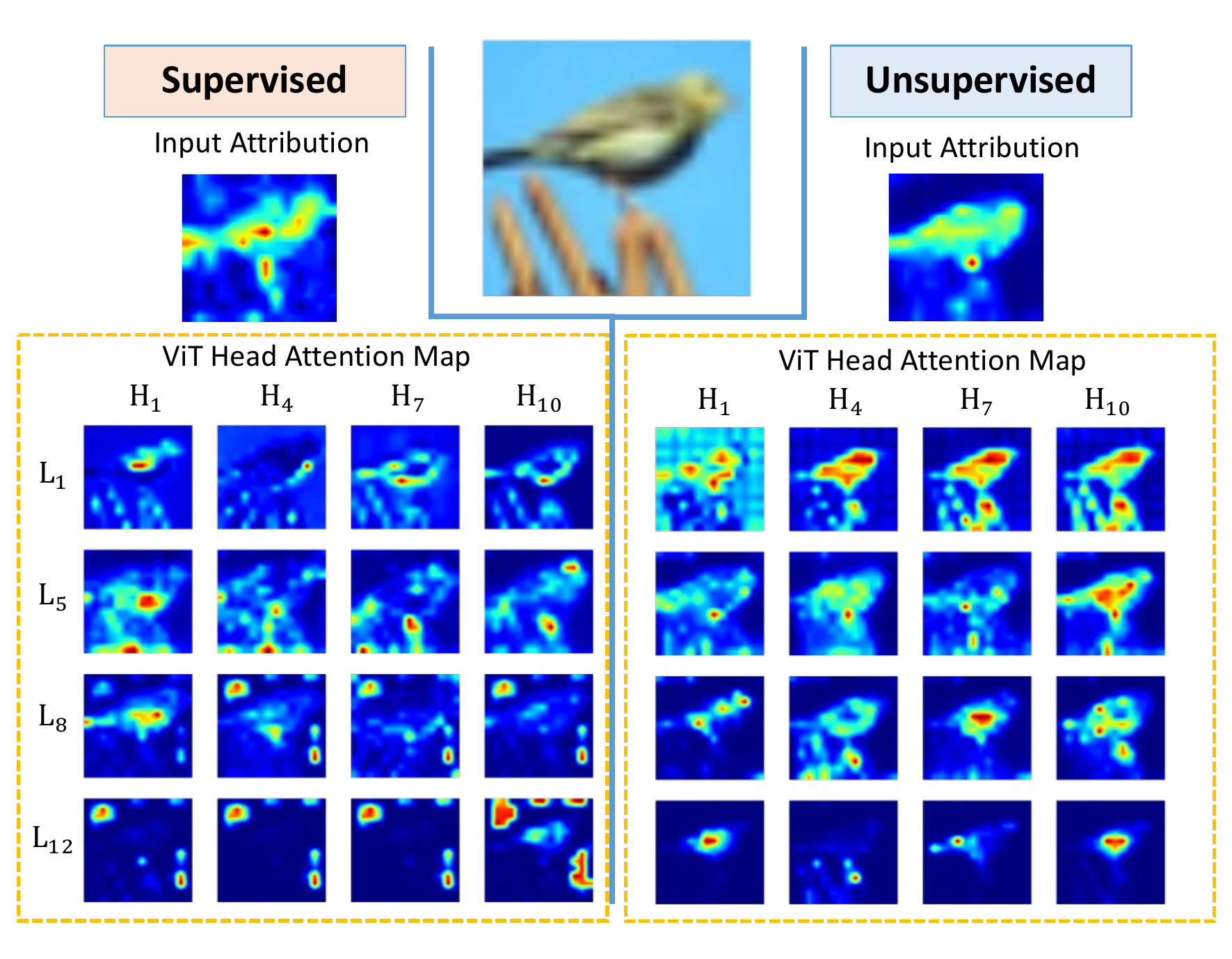}
\caption{SmoothGrad input-attribution and Attention maps for heads in some layers. We use ViT-ImageNet21K\cite{imagenet21k}  for the supervised model, and ViT-BEiT\cite{beit} for the unsupervised model. Both of the input-attribution highlights the object region, while the attention maps are clearly different.}
\label{1_introduction}
\end{figure}

% Pretraining 
In general, there are  two ways of pretraining: supervised and unsupervised pre-training \cite{khan2022transformers_survey}. In the supervised pretraining, ViT is trained to predict the label of an image with a classifier branch. 
On the other hand, the unsupervised pretraining learns the data patterns without labels. 
Both of them are known to be effective for downstream tasks and in general they show similar generalization performance.  
Therefore, both of them could be treated equally in the perspective of generalization, but the internal mechanisms are still undiscovered. For example, Figure \ref{1_introduction} shows the input-attribution and the attention maps of the finetuned ViT on image classification. Two pretrained models have different weights on the critical region of the image.  Even though MSA directly gives information on the important image patches, to the best of our knowledge, there is no quantitative method for all the heads in the transformer. Therefore, concrete analysis and comparison on different transformer architecture are challenging. 

To alleviate the burden of analysis, we propose \textbf{I}nput Attribution and  \textbf{A}ttention  \textbf{S}core (IA-Score), a quantitative metric for attention map by computing the agreement with input-attribution method such as GradCAM \cite{selvaraju2017gradcam} and SmoothGrad \cite{smoothgrad}. In addition, we propose \textbf{IA}-Score \textbf{V}ector (IAV), a concatenation of IA-score of all heads in ViT encoder layers. Visualizing IAV  directly gives an insight on the role of heads and eases the comparison on  pretrained ViT models.

We empirically show that (1) IAV can explain the pretraining types of the fine-tuned model (supervised or unsupervised) 
and (2) IAV is somehow global property of ViT and not a input specifically different.

\begin{figure*}[t]
\centering
\includegraphics[width=15.5cm]{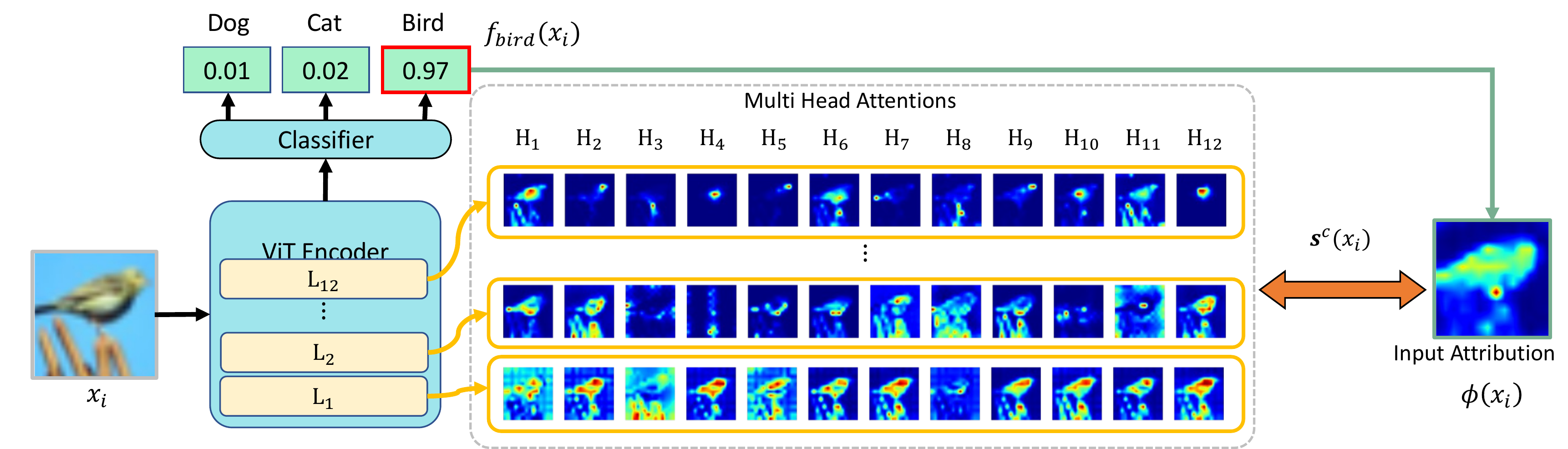}
\caption{ We obtain input-size heatmaps, input-attribution $\phi(x_i)$ which explains the important input regions for class decision and attention map $A_{l,h}(x_i)$ which explains the internal mechanism of ViT. Each head is evaluated by the similarity score with the input-attribution. There are total $12 \times 12 =144$ heads in the ViT Encoder. All the images are obtained with the finetuned BEiT on CIFAR 10.  }
\label{2_framework}
\end{figure*}

\section{Related Work}

Multi-Head Attention is widely studied for various reasons. For Interpretation, information passing based attention rollout is proposed \cite{attention_flow}. 
In Natural Language Processing, various analysis includes the role of head  \cite{clark2019does_bert_attn}, task-specific difference of attention maps \cite{transformer_attn_task_specific}, and part of speech \cite{baan2019understanding_grammer, vig2019analyzing_language_structure}. Therefore, the domain specific properties of multi-heads are discovered. In computer vision, DINO \cite{dino} showed the attention patterns of pretrained model, but did not analyzed each head. 

Attention patterns further be used for regularization such as  attention pruning \cite{voita2019analyzing_attn_prune}, disagreement over attention maps \cite{li2018multi_attn_regularization}, reduce similarity over layers \cite{zhou2021deepvit_attn_similar} , reduce redundancy \cite{kim2021rethinking_attn_sparse}, and remove ambiguity using decision tree \cite{attn_tree_ambiguity}. The previous studies mainly focused on the property of attention maps. 
To the best of our knowledge, our work is the first one to study on the internal representation of pretraining bias by quantitatively using attention maps. 

\section{Method}

\subsection{input-attribution and Attention Score}
Consider neural network $f:\mathbb{R}^{W\times H\times C} \rightarrow \mathbb{C}$ with width $W$, height $H$, channels $C$, and class set $\mathbb{C}$. The decision score on class $c\in \mathbb{C}$ for  image $x_i$ is denoted by $f^c(x_i)$ and the input-attribution on $x_i$ is $\phi^c(x_i)$. As we consider ViT architecture, the attention score of class token over image patches is denoted by  $A_{l,h}(x_i)$. The similarity of input-attribution and the attention score of a single head is defined by
\begin{equation}
s_{l,h}^c(x_i) = \cos{\Big(\phi^c(x_i), A_{l,h}(x_i)  \Big)}
\end{equation}
where $\cos(\cdot, \cdot)$ is the cosine similarity between two vectors. We call $s_{l,h}^c$ \textbf{I}nput-attribution and \textbf{A}ttention \textbf{Score} (IA-Score) as it reduces attention map to a quantitative score. 
Note that the input-attribution is pixel level metric while the attention score is patch level. To match the dimension, we average the input-attribution value of pixels in a patch to make the same size. In addition, we assume non-negative input-attribution and monotonically increasing importance from 0 to 1. The score is normalized to guarantee, that is,  $||\phi^c(x_i)||_2 = 1$ and $||A_{l,h}(x_i)||=1$. Therefore, the scale of input-attribution has the same meaning with the attention score. 

One important observation is that the attention score is conditional only on the input, while the input-attribution is conditional on both input and target class. Therefore, IA-Score shows the agreement between the ViT attention and the class-specific input-attribution. 

\subsection{input-attribution and Attention Score Vector}
For interpretation of the overall ViT, the input-attribution $\phi^c(x_i)$ is paired with all heads in the ViT. For $H$-heads and $L$-layers ViT, we concatenate $s_{h,l}^c(x_i)$ for $h \in \{1,\cdots, H\} \text{~and~} l \in \{1,\cdots, L\}$ and  obtain $H\times L$ dimensional vector, called \textbf{IA}-Score \textbf{V}ector (IAV)
\begin{equation}
\mathbf{s}^c(x_i) = [s_{1,1}^c(x_i), \cdots, s_{L,H}^c(x_i)]^{\mathrm{T}}
\end{equation}
IAV is a structured vector as heads in a transformer layer are parallel and the layers are sequential. Therefore, even though IAV is vector representation, the analysis must consider the structure of heads and layers. Figure \ref{2_framework} shows the overall structure.

Even though IAV is directly dives the vector to evaluate the attention heads, it is still hard to analyze the global property of the ViT. We present  global-IAV which is an average of IAV over samples. 
\begin{equation}
\mathbf{s}_f = \sum_{i}^N \mathbf{s}^{\hat{y}}(x_i) 
\end{equation}
where $\hat{y}$ is either a predicted or a ground truth label. 
$\mathbf{s}_f$ represents the global role of heads and each element could be interpreted by the agreement of input-attribution and the attention map. As well as the input-attribution, the similarity score could be computed with any attributions whose size is same with the input. For example, we could measure the global receptive ratio by just replace $\phi^c(x_i)$ with the segmentation map of an object. We generalize the IAV with any input type and call it Any-AV(AAV) .

\begin{figure*}[ht]
\centering
\includegraphics[width=17cm]{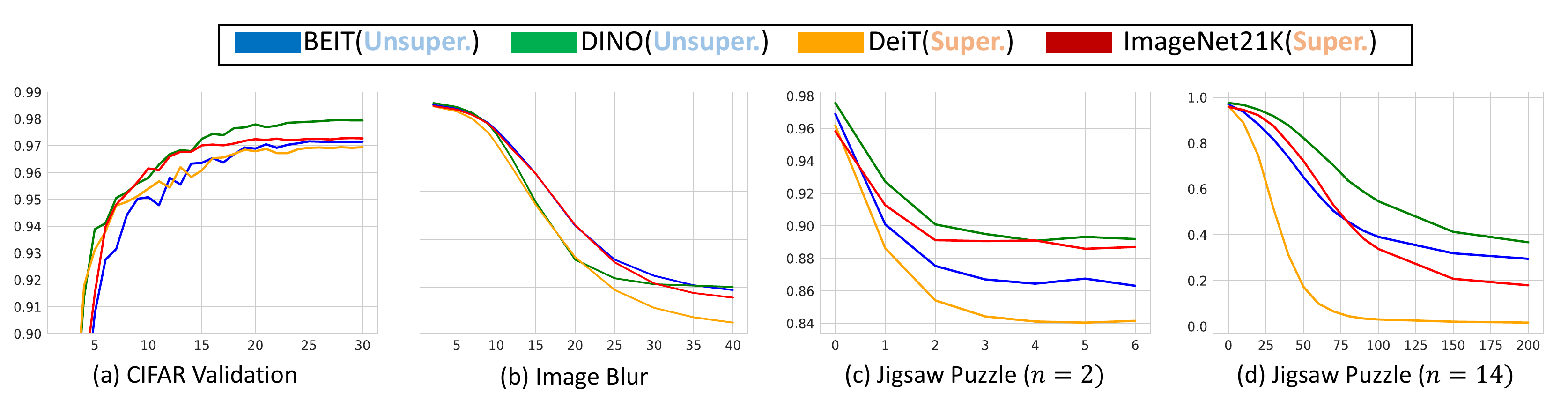}
\caption{Evaluation accuracy on the finetuned models. (a) CIFAR 10 validation accuracy over epochs. (b) blurred dataset accuracy with blur degree on x-axis. (c) jigsaw puzzle with grid size 2 and the number of swaps on x-axis. (d) jigsaw puzzle with grid size 14 which is the number of patches in ViT. The performance of DeiT dropped faster than other models in robustness test. DINO showed the best generalization and robustness on the data distribution shift. The generalization and the robustness are not characteristic of unsupervised pretraining as ImageNet21K also showed similar performance and BEiT showed not a significant performance like DINO. }
\label{3_eval}
\end{figure*}

\section{Experiments} \label{Experiments}

We divide pre-training types into (1) unsupervised and (2) supervised pretrainings. In the case of unsupervised, Bidirectional Encoder representation from Image Transformers (BEiT) \cite{beit} and DINO \cite{dino} are used. In the case of supervised, DeiT \cite{deit} ,which is trained only on ImageNet1K, and ViT on pretrained on ImageNet21K are used. We further finetuned the models with CIFAR 10 dataset \cite{krizhevsky2009learning_cifar10} which has 50K images in total. We use 768 batch size, 1e-5 learning rate with consine annealing schedule, and 30 epochs which is enough for the validation to converge. 
After the models are finetuned, we evaluate the models to obtain proper comparisons between supervised and unsupervised pretrained models. We summarize the \textbf{target analysis} and the treatment on it as follows
\begin{enumerate}[label=\protect\circled{\arabic*},  topsep=1.4pt, before=\setstretch{1.05}]
\item \textbf{Generalization and Robustness}: we evaluate the model performance with validation data for generalization, For robustness, we shift the data distribution in two ways: the image blur and jigsaw puzzle. In jigsaw puzzle the image patches are shuffled.  
\item \textbf{Dynamics of Attention and input-attribution}: we measure how much change occurs in training for input-attribution and attention maps. We evaluate the difference with the final model and intermediate models. 
\item \textbf{Importance of Attention Maps}: we mask a portion of input image based on the attention score on the patches to see whether the heatmaps obtained by attention is meaningful or not. We compare it with other methods.
\item \textbf{Global Receptive Field}: we measure the entropy of the attention map for each head for a single sample and average it over all the samples to see the spread degree of attentions. 
\item \textbf{global-IAV and global-AAV} : we compute IAV with the finetuned models and compare the global-IAV for supervised and unsupervised models. In the case of AAV, we use SmoothGrad of DINO as it attends highly on the global input region. 
\item \textbf{The Role of Heads}: We present the boxplot of IAV over all samples and compare the value intervals so that the interval of IA-Score could be visualized. We empirically show that there are two types of heads: High-IA-Head and Low-IA-Head.
\item \textbf{Class Discrimination}: We present t-SNE manifold with IAV.  We compare the manifold layer by layer so that the IAV can be used to discriminate the classes. 
\end{enumerate}

\section{Results}

In this section, we present the empirical results on the analysis targets in Section \ref{Experiments}. Each subsection is aligned to the indexing of the analysis target. 

\subsection{Generalization and Robustness}\label{g_and_r}
Figure \ref{3_eval} shows the evaluation results on the finetuned models. In the case of robustness, we report the performance with the final checkpoint. Figure \ref{3_eval} (a) shows   validation accuracy for four models. We found that DINO had the superior performance with our training setting ($98\%$). However, other models also showed competitive performance ($\approx 97\%$) and BEiT showed similar performance with other supervised models. Therefore, there is no significant meaning on the comparison of pretraining methods by performance. Even though DINO showed the better performance, other models could achieve similar performance with proper regularization. 

For the robustness, we report the performance by having more data distribution shift. We tested robustness in two ways: (1) image blur and (2) jigsaw puzzle which swaps the image patches. Figure \ref{3_eval} (b) to (d) shows the CIFAR 10 evaluation accuracy with distribution shifts. We found that DeiT is significantly weak at data distribution shift, while other models show competitive performance. ImageNet21K model showed similar trends in both ImageBlur and Jigsaw  Puzzle evaluations. By considering that the ImageNet21K is trained on more large dataset, we conclude that the robustness could be achieved by the amount of training data, instead of pretraining methods. In conclusion, the comparison of supervised and unsupervised models by validation set and data distribution shift is not capable.  

\subsection{Training Dynamics}
\begin{figure}[ht]
\centering
\includegraphics[width=7.5cm]{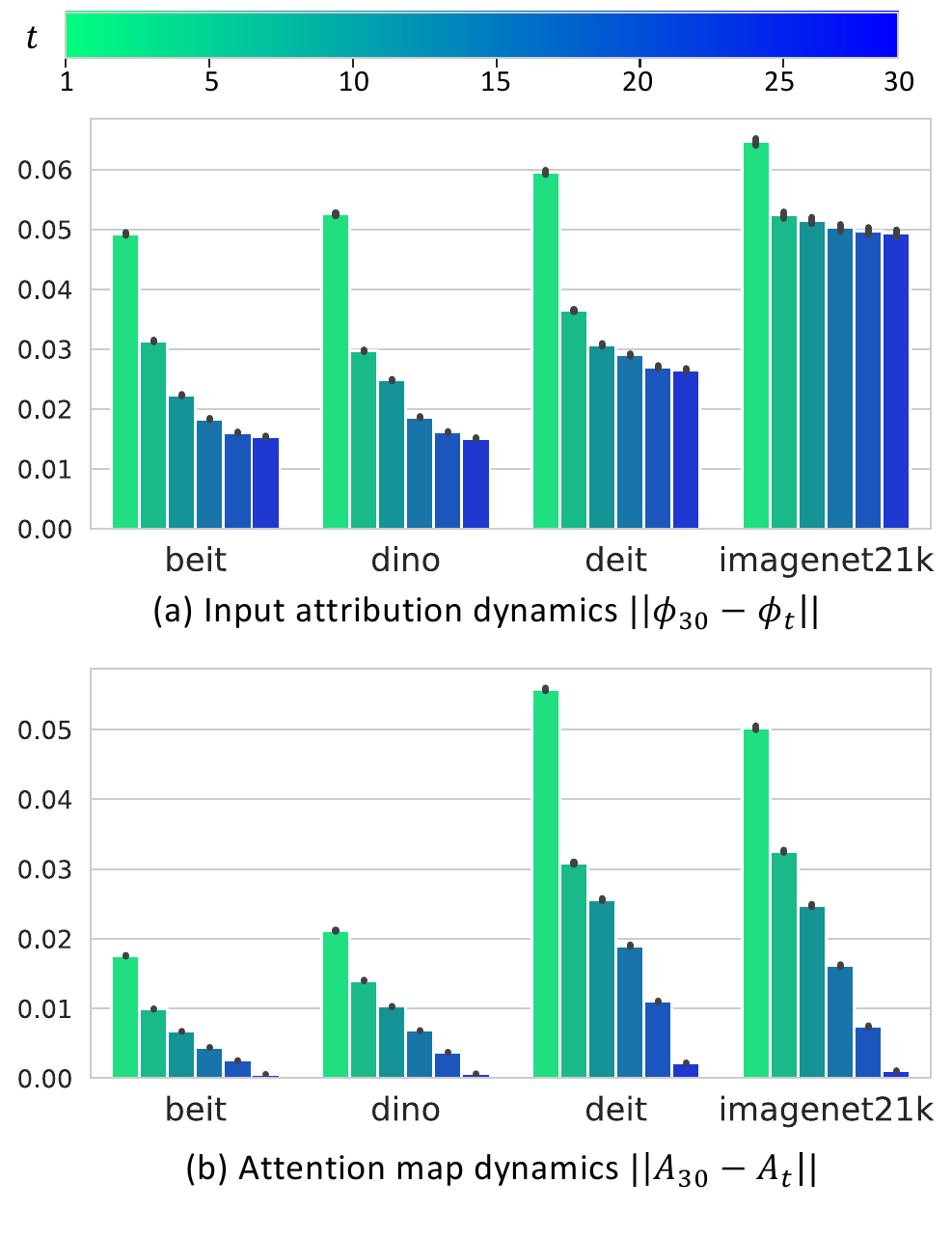}
\caption{
The difference of the heatmap for checkpoint models and the final model. (a) is SmoothGrad difference over samples and (b) is the attention map averaged over all layers. Unsupervised models have less magnitude on input-attribution and significantly lower magnitude in attention map.  
% Supervised Model has large deviation from the 
}
\label{4_dynamics}
\end{figure}

We measured the difference of attributions  for the trained models. Figure \ref{4_dynamics} shows the difference between training epoch $t$ and the final epoch $30$. We use notation $\phi_t$ for SmoothGrad and $A_t$ for attention map. 

As the figure shows, there is a difference ratio of the change by the pretraining methods. Both  input-attribution and attention maps are changed less in  unsupervised models (BEiT and DINO). Note that supervised models (DeiT and ImageNet21K) had similar magnitude (0.05) for both input-attribution and attention map, while unsupervised models changed almost half (0.02). The result indicates that unsupervised models have less change on the attention map while training. 

Note that ImageNew21K showed similar robustness on the robustness evaluation, yet the internal attention map and the input attribution changes more dramatically. 
The result shows that the supervised pretraining has more dynamical changes in the attention map than unsupervised models.

\subsection{Input Disturbance}
\begin{figure}[ht!]
\centering
\includegraphics[width=7.6cm]{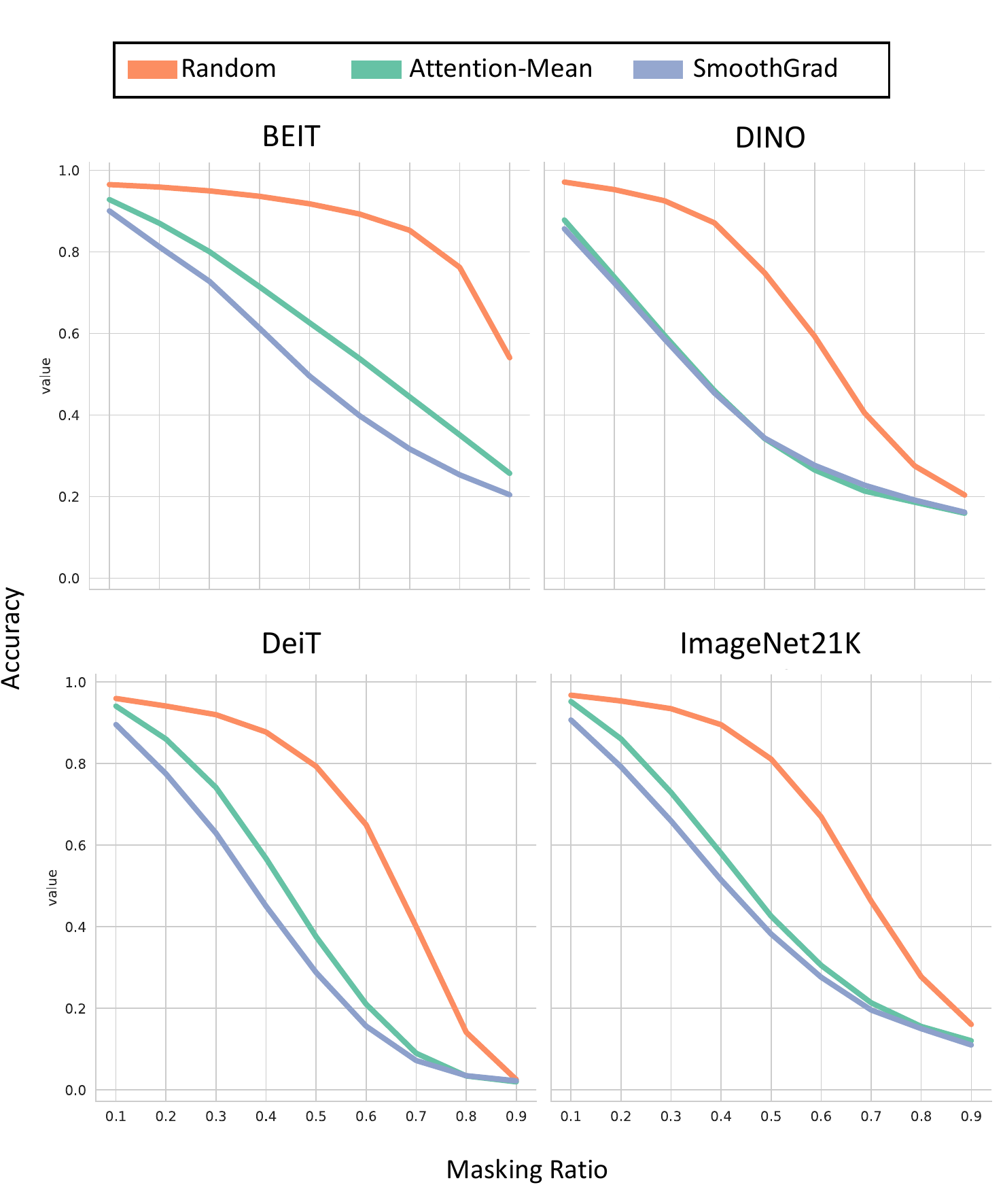}
\caption{Validation accuracy on the masked inputs. x-axis is the masking ratio from 0.1 to 0.9. Only Attention map of DINO showed the similar performance drop with SmoothGrad.  
As the gap between SmoothGrad and other methods is large for BEiT, we could not conclude that attention map itself effectively explains the decision of the model.  
}
\label{6_filter_input}
\end{figure}

As we discussed earlier, there is a believe that the attention map presents important region of model decision. We give the result on the believe by masking the regions of the attention map. We also present the results of SmoothGrad and random based selection for comparison. Figure \ref{6_filter_input} shows the evaluation accuracy on CIFAR 10 validation set over masking ratio. Attention-Mean is the mean pooling over all heads in layers. 

We empirically found that only DINO had similar performance with SmoothGrad and other models show that SmoothGrad has more effects on the class decision. 
One reason would be the property of DINO. DINO is known to make attention maps on the input region for the pretrained model, while other models focus on other regions. 
In addition, as Attention-Mean of BEiT had different results with DINO, we could not conclude that the effect of attention map on the class decision meaningfully separate the pretraining methods.

\begin{figure*}[ht]
\centering
\includegraphics[width=16.5cm]{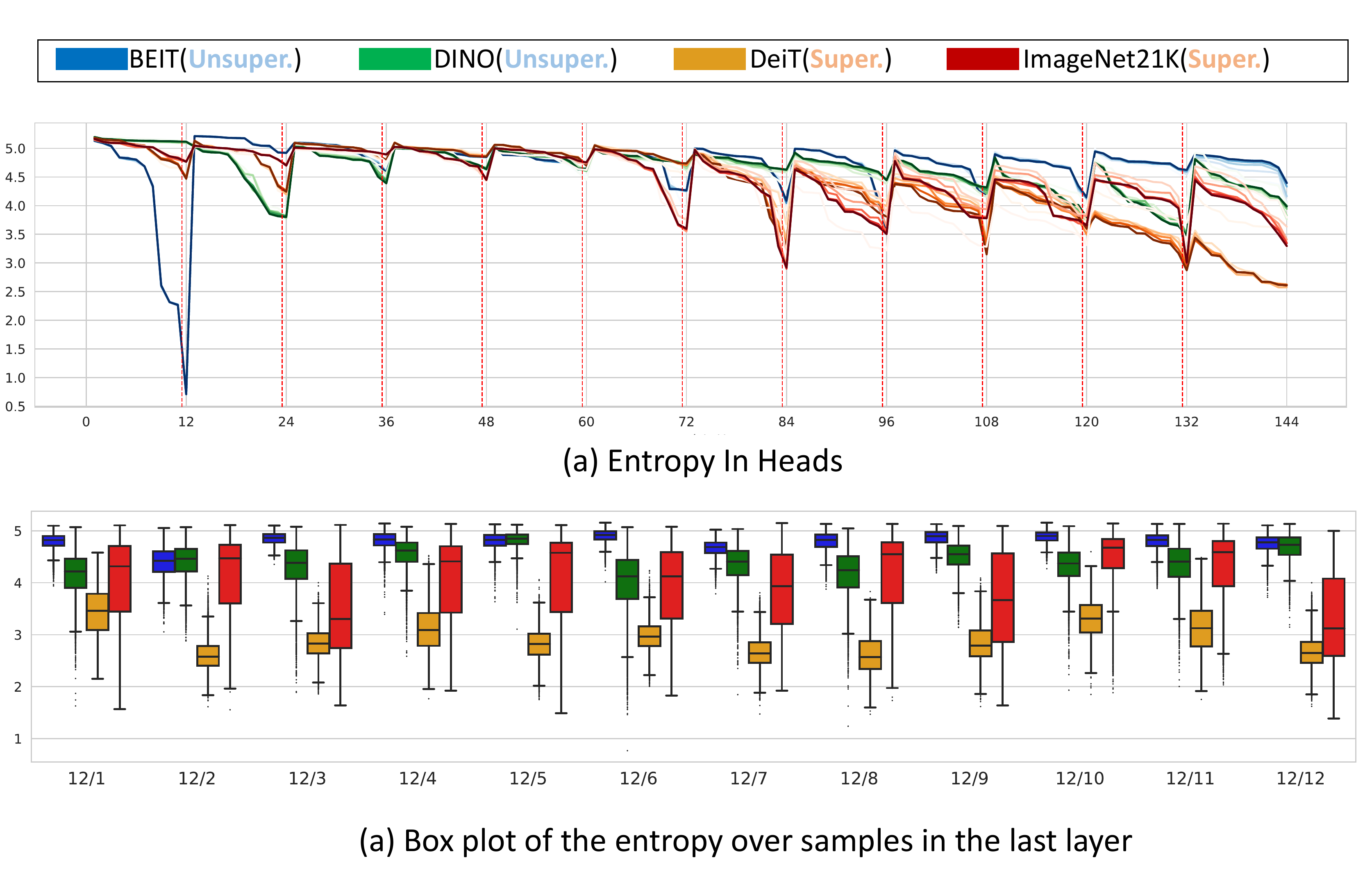}
\caption{The entropy of the attention maps for supervised and unsupervised models.  (a) the dynamics of entropy in training. The darker color indicates more training epochs. We report the average of entropy over all heads. We could not conclude that higher entropy is the property of unsupervised model as unsupervised models show generally high entropy at the end of the layers, while supervised models  have higher entropy in the early layers. (b) the box plot of entropy in the last layer $L12$ over all samples. Unsupervised models show high entropy while supervised models show low entropy. However, ImageNet21K show similar entropy for some heads with DINO model. Therefore, we could not say supervised pretraining yields low entropy.  }
\label{5_attn_entropy}
\end{figure*}

\subsection{Global Receptive Field}

As the transformer has attention maps, we can directly compare the models  with a quantitatively measure on attention maps. For quantitative measure, we use entropy of an attention map which represents how much the attention scores are spread. We compute the entropy on a single head and averaged it  over all samples to measure the scope of entropy. 
\begin{equation}
    \sum_i^N H(A_{l,h})
\end{equation}
where $H(A) = -\sum_p A(p) \log A(p)$ is the entropy of attention score $A(p)$ on pixel $p$ (or patch). Figure \ref{5_attn_entropy} (a) shows the dynamics of entropy in training for 30 epochs for four models. We sorted the heads in the same layer as they are processed in parallel. 

In general, the supervised models show increasing trend, while unsupervised models show decreasing trend in training. Therefore, attention maps are differently biased based on the pretraining types in dynamics. However, we could not distinguish pretraining types by the entropy score when it is fully trained as DINO and ImageNet21K had similar degree on the layers $L1$ to $L11$. 

We observe that the final layer $L12$ has a meaningful difference and we present the box plot of entropy over all samples in Figure \ref{5_attn_entropy} (b). 
 BEiT and DINO have relatively large entropy with low variance and DeiT has significantly lower entropy with low variance. Interestingly, ImageNet21K has some heads whose median is similar with DINO and we could not conclude that supervised pretraining always makes low entropy heads.  Similar to the conclusion of robustness in Subsection \ref{g_and_r}, one reason of high entropy for ImageNet21K could be the large amount data in training. 
 
Although entropy analysis has meaningful comparison on pretraining methods, there is a critical limitation on the entropy analysis. Entropy does not consider the relative object size in the input image. The entropy could be low if the object occupies small region such as a basketball ball detection. Therefore, even though the attention map properly has high density on object, we obtain low entropy. 

Unlike the entropy measure which does not consider the ciritical region of the input, IAV does not suffer this problem because a baseline is given and it has high value only on the critical region. In addition, IAV insures a fair comparison between all the heads because the same baseline used for all heads.

\begin{figure*}[ht]
\centering
\includegraphics[width=16cm]{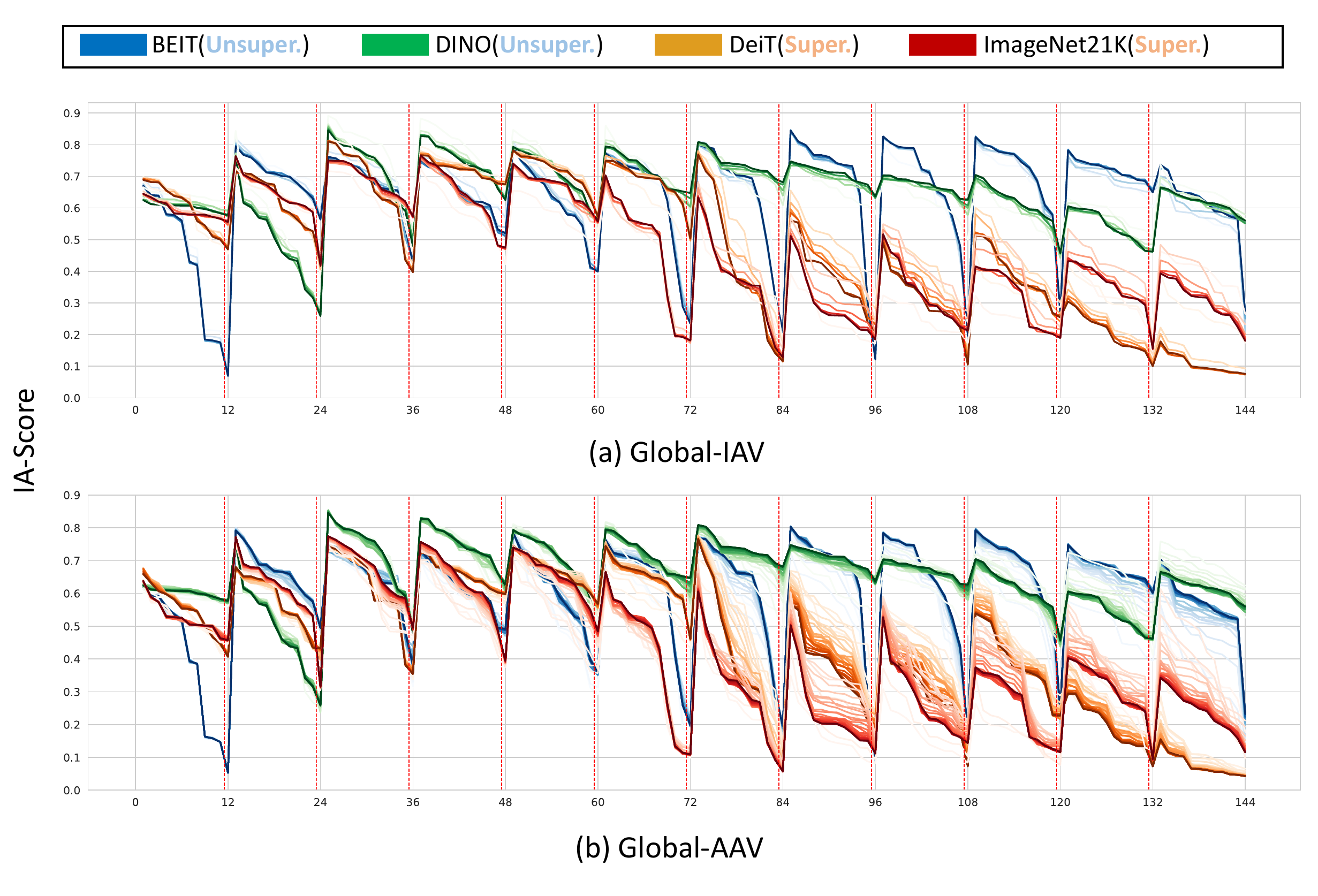}
\caption{global-IAV and global-AAV training dynamics. The darker color indicates more training epochs. the scores are averaged over samples and the heads in the same layer are sorted in decreasing order. (a) global-IAV which computes the IA-Score with attention map and SmoothGrad input-attribution. Unsupervised models have higher IA-Score after 6th layer, while supervised  models have decreasing trend.  (b) global-AAV with SmoothGrad of the last checkpoint DINO model which represents the global receptive field of unsupervised models. Supervised models have significant decreasing trend, while BEiT shows similar dynamics.  }
\label{7_iav}
\end{figure*}

\subsection{Global-IAV and Global-AAV}

Unlike the entropy analysis which produce a quantity only with an attention map, IA-Score is measured with critical region of an input image. 
One advantage is that IA-Score offers a common baseline to heads. Because we compute the similarity score with a baseline, 
the value could be understood in more intuitive manner. 
As the score is close to 1, the attention map of a single head could be considered to be agreed with the model decision, while close to 0 means disagree with the model decision. 

Figure \ref{7_iav} (a) shows the global-IAV with SmoothGrad input-attribution in training models.  
The global-IAV trend of supervised model is  decreasing, while the trend of unsupervised model is stable (not decreasing). 
One reason of such decreasing trend is that the heads of supervised model is biased to focus on  independent region of the input. 
On the other hand, the heads of later layers in the unsupervised models are still focusing on the input regions.

As commented before, the agreement of attention maps could be measured with any type of input. 
Figure \ref{7_iav} (b) shows global-AAI with SmoothGrad of the final checkpoint DINO. 
In this case, the IA-Scores with supervised models are reduced much more than global-IAV in Figure \ref{7_iav}  (a). One of the future research could be done with various types of input types such as segmentation. We leave this study for further application of AAV.

One important observation is that the IA-Scores are not equally distributed in a single layer, but shows a specific interval.
In the case of ImageNet21K and DeiT, the middle layer has the most wide spectrum. 
The spectrum represents that each head has a different degree of agreement with input-attribution and supervised models have much less agreement than unsupervised models. 
We found that BEiT and DINO have high IA-Score, while there are 1 or 2 heads in BEiT whose IA-Score is significantly lower. 
Therefore, the most of heads in BEiT are similar with DINO, yet there are few heads whose behavior is different.

Based on the observations that there is a large IA-Score gap between transformer heads,  we weakly propose a general property of the pretrained ViT. The attentions on important region of input are not equally distributed over heads and it is not sample dependent property and rather head-dependent. Therefore two types of heads exists.  We provide further supports on the conjecture in the next section.  

\begin{enumerate}[label=\protect\circled{\arabic*},  topsep=1.0pt, before=\setstretch{0.9}]
\item Some heads are apathetic on input region
\item Some heads are enthusiastic on input region
\end{enumerate}

\begin{figure*}[ht]
\centering
\includegraphics[width=16cm]{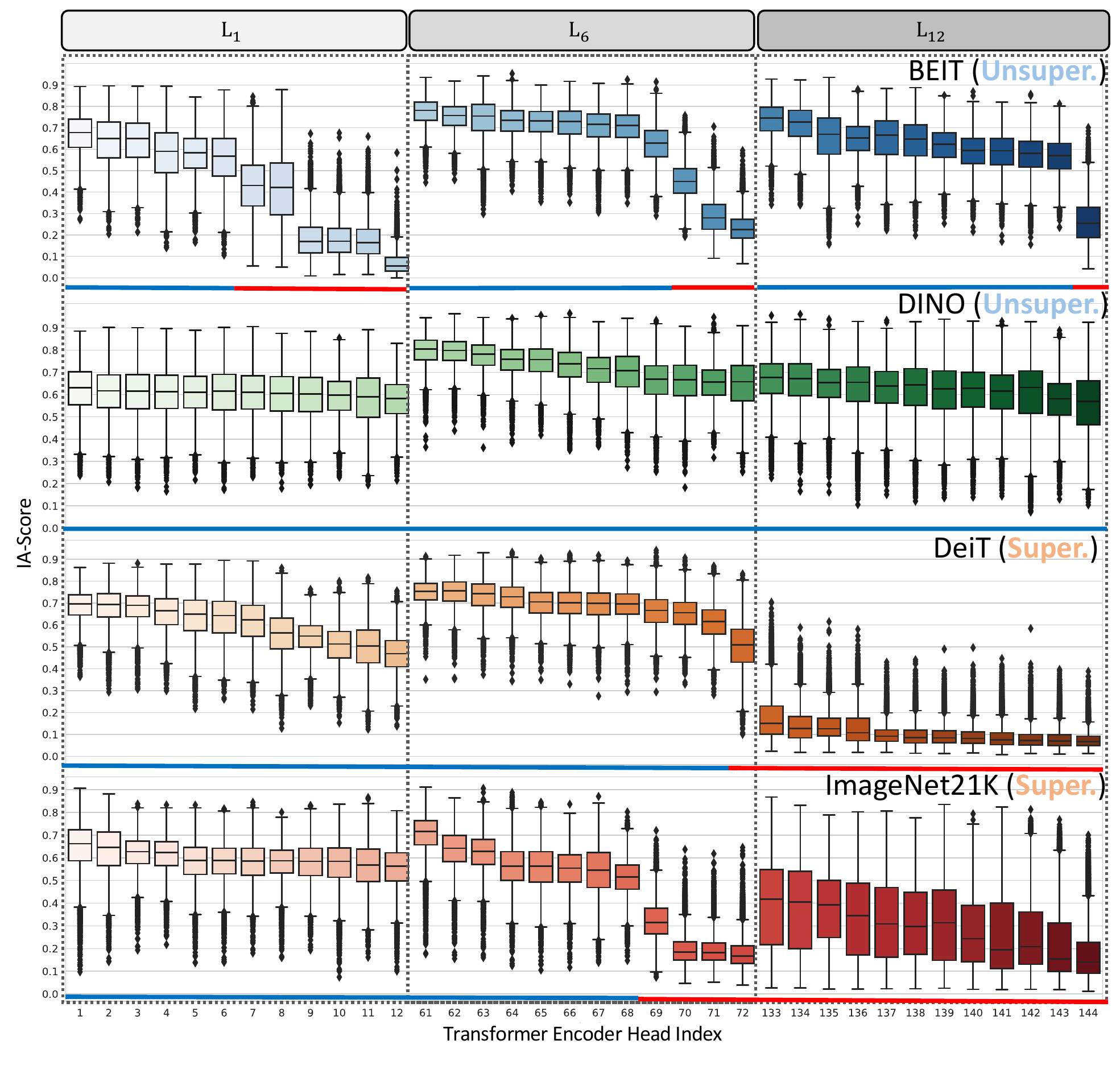}
\caption{Underline color shows heads with larger than 0.5 (\textcolor{blue}{blue}) and smaller than 0.5 (\textcolor{red}{red}) IA-Score. DINO has only high IA-Score heads, and BEiT have high and low IA-Score heads over layers. Supervised pretrianing models have high IA-Score heads in early layers, while low IA-Score heads exists on the later layers.  }
\label{8_iav_var}
\end{figure*}

\subsection{The Role of Heads}

To validate the conjecture given in the previous section, we analyze the IA-Score in detail. 
We divide heads into two types so that we can compare pretraining methods in convenience manner:  
\begin{itemize}
    \item \textcolor{blue}{High-IA-Head} :  median IA-score is larger than 0.5
    \item \textcolor{red}{Low-IA-Head} :  median IA-Score is smaller than 0.5
\end{itemize}
Figure \ref{8_iav_var} shows IA-Score of ViT heads over all CIFAR 10 validation samples.
We marked IA-Head types with colors in the below lines of each models. Because of the limitation on the visualization, we present only the first, middle, and the last layer of the ViT. We acknowledge that the head types further be divided by the variance of IA-Score. We leave a deeper study on head types for future study. 

Unsupervised models  mostly have High-IA-Head over all layers and the AI-Score preserved along the layers. In addition, the general trends of three layers ($L1, L6, \text{~and~} L12$) are similar.  In the case of DINO, only High-IA-Head type exists, and BEiT has dominantly High-IA-Heads but there are a few Low-IA-Heads. We guess that Low-IA-Head of BEiT have similar role with Low-IA-Heads in supervised models. 

Unlike the unsupervised pretraining, supervised models have High-IA-Heads in early layers and Low-IA-Heads on deeper layers. Therefore, two types of heads exists in the supervised models. 
DeiT has High-IA-Heads in the early layer, a few Low-IA-Heads in the middle layer, and only Low-IA-Heads in the last layer. 
ImageNet21K has similar trend with DeiT but has Low-IA-Heads with higher variance.

\begin{figure*}[ht]
\centering
\includegraphics[width=17.0cm]{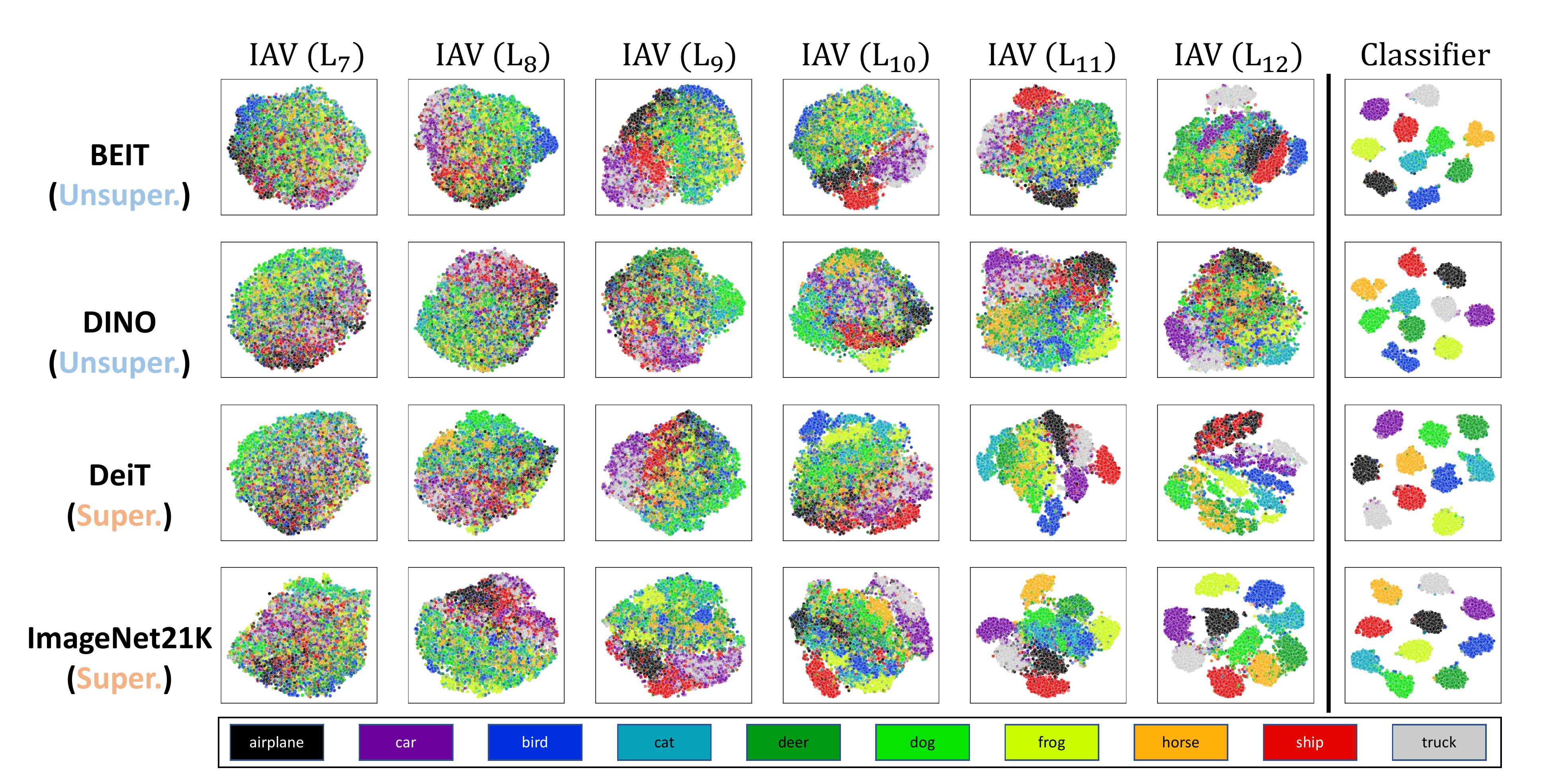}
\caption{Classes can be grouped by IAV patterns on both supervised and unsupervised pretraining. 
As the layers are deeper, the separation by IAV patterns is more clear. Supervised models shows even more clear distinction at layers $L11$ and $L12$. DeiT shows entangled IAV patterns than ImageNet21K which explains why DeiT shows worse performance on evaluation. Unsupervised training shows relatively entangled representations. BEiT shows better disentanglement than DINO. }
\label{9_tsne}
\end{figure*}

\subsection{Class Discrimination}
For the last result, we show that IAV, which measures the agreements,  itself has explanation. We found that IAV patterns can be used to classify the labels and the trend is more clear on supervised models. Remind that IAV is just a measure on the heads to calculate the agreement score to the input-attribution and does not contain any information on the input data. 
    
Figure \ref{9_tsne} shows the t-SNE \cite{van2008visualizing} plot of IAV for each layer (12 dimensional vector for each layer) and the classifier of each model. The result shows that the IAV pattern successfully classify the input as the layers are deeper. The supervised pretraining has more clear clusters than unsupervised pretraining. 
Therefore, the result also supports that the main idea that IAV separates the pretraining types.

We provide a guess on why the clusters are formed with IAV. Previously, it is known that the classification is done based on the attention of input region and it is not all the mechanisms of transformer. Additional classification is done with the head agreement patterns. 
Each head has different degree on the agreement, and therefore, the pattern of the agreement over heads makes the classification possible. In addition, this trend is more clear for supervised models.

One of the core contribution of t-SNE IAV plot is the interpretation of internal layers for the transformer. Previously only the attention map is available for the analysis. However, we can explain the internal mechanism with IAV and importantly we can compare the finetuned pretrained models with the plot.

\section{Conclusion}

Pretraining is essential for large scale models as it reduces training time and generalizes well. 
Supervised pretraining learns to predict a label and the internal representation can distinct the labels of data well. 
On the other hand, unsupervised pretraining learns the patterns in the data distribution and the internal representations can be used cluster data patterns. 
Both of pretraining types are helpful for finetuning on down stream tasks, yet the internal representations are quite different. 
To explain the internal representation of ViT with attention maps, we propose methods to quantitatively analyze them. 
We propose IA-Score which measures the agreement between attention map of an ViT head and the critical regions of the model decision called input-attribution. 
In addition, we propose IAV which shows IA-Score pattern over all heads in ViT. We empirically show that IAV separates classes and the discrimination of classes is more clear for supervised pretraining. 
Even though there is a variance of IAV patterns over all samples, the global-IAV trend can distinguish the pretraining types. 

We expect IAV to be used in various fields where the explanation of finetuned ViT is particularly important. 
As we provide t-SNE of IAV patterns on layers, such analysis could be used to understand the role of each layer in ViT. 
One limitation is that we do not provide more detailed explanation why the IAV patterns are shaped differently and why there are low IA-score heads exist and role of them. 
In future, we will study on the property of individual heads.

\clearpage
\bibliography{references}

\end{document}